\documentclass[sigconf]{acmart}
\usepackage{subcaption}
\AtBeginDocument{%
  }

\copyrightyear{2024}
\acmYear{2024}
\setcopyright{rightsretained}
\acmConference[CHI '24]{Proceedings of the CHI Conference on Human Factors in Computing Systems}{May 11--16, 2024}{Honolulu, HI, USA}
\acmBooktitle{Proceedings of the CHI Conference on Human Factors in Computing Systems (CHI '24), May 11--16, 2024, Honolulu, HI, USA}
\acmDOI{10.1145/3613904.3642459}
\acmISBN{979-8-4007-0330-0/24/05}

\newcommand*{\rr}{\textcolor{black}}
\begin{document}

\title[Generative Echo Chamber?]{Generative Echo Chamber? Effects of LLM-Powered Search Systems on Diverse Information Seeking}

\graphicspath{{figures/}}

\author{Nikhil Sharma}
\email{nsharm27@jhu.edu}
\affiliation{%
  \institution{Johns Hopkins University}
  \streetaddress{3400 N Charles St}
  \city{Baltimore}
  \state{Maryland}
  \country{USA}
  \postcode{21218}
}

\author{Q. Vera Liao}
\email{veraliao@microsoft.com}
\affiliation{%
  \institution{Microsoft Research}
  \city{Montr\'eal}
  \country{Canada}}
  
\author{Ziang Xiao}
\email{ziang.xiao@jhu.edu}
\affiliation{%
  \institution{Johns Hopkins University}
  \streetaddress{3400 N Charles St}
  \city{Baltimore}
  \state{Maryland}
  \country{USA}
  \postcode{21218}
}

\renewcommand{\shortauthors}{Sharma et al.}

\begin{abstract}

Large language models (LLMs) powered conversational search systems have already been used by hundreds of millions of people, and are believed to bring many benefits over conventional search. However, while decades of research and public discourse interrogated the risk of search systems in increasing selective exposure and creating echo chambers---limiting exposure to diverse opinions and leading to opinion polarization, little is known about such a risk of LLM-powered conversational search. We conduct two experiments to investigate: 1) whether and how LLM-powered conversational search increases selective exposure compared to conventional search; 2) whether and how LLMs with opinion biases that either reinforce or challenge the user's view change the effect. Overall, we found that participants engaged in more biased information querying with LLM-powered conversational search, and an opinionated LLM reinforcing their views exacerbated this bias. These results present critical implications for the development of LLMs and conversational search systems, and the policy governing these technologies. 

\end{abstract}



\maketitle

\section{Introduction}
Exposure to diverse viewpoints is essential for critical thinking, balanced views, and informed decision-making, and at a collective level, preventing opinion polarization or even dangerous radicalization. However, such ideals are often challenging to achieve because people have a natural tendency of selective exposure~\cite{frey1986recent}, or confirmation bias~\cite{wason1960failure}, favoring consonant information and avoiding dissonant information. In the last two decades, much research and public discourse have expressed concerns regarding the exacerbating effect of information and web technologies on selective exposure. For example, by personalization and supplying only information people want to see, search engines and recommender systems such as news feeds may produce ``filter bubbles''~\cite{pariser2011filter} of ideological and cultural isolation. By allowing people to easily select whom to connect with and what they want to see, social media can create ``echo chambers'' where people end up only interacting with like-minded others. In short, information technologies can have multiplex mechanisms to exacerbate people's selective exposure bias, from data and algorithmic biases, to biases induced by interaction affordances~\cite{baeza2020bias}.  

With the recent rise of powerful large language models (LLMs) such as the GPT-4, a new generation of LLM-powered information technologies has emerged, from conversational search, open-domain or specialized chatbots, to productivity tools such as writing support. These technologies can have profound effects on the information consumption of individuals and the society at large. First, LLMs are in essence ``next token predictors'' that optimize for giving expected outputs, and thus can potentially be more inclined to provide consonant information than traditional information system algorithms. However, LLMs can be used to provide ``synthesized'' content (e.g., a concise summary) based on a collection of documents, which may help expose people to diverse information by removing biases that they may exhibit when selecting which document to read. Information technologies powered by LLMs also support natural and interactive conversational interactions. How these new affordances of interaction shape people's information-seeking behaviors remains an open question. Last but not least, LLMs are known to encode biases from the training data~\cite{rae2021scaling,abid2021persistent}, and can be easily steered to exhibit certain opinion biases through model adaptation techniques such as fine-tuning or prompting~\cite{bommasani2021opportunities}. Little is known whether the encoded opinion biases can exacerbate selective exposure of people with similar views, or be used to expose people to different viewpoints. Given the rapidly growing reach and usage frequency of LLM-powered information technologies, it is paramount for the research community to investigate these issues and LLMs' effect on information diversity to inform the development of LLMs, design of LLM-powered systems, as well as policy governing these technologies. 

In this work, we take a formative step toward understanding the echo chamber effect of LLMs by focusing on LLM-powered conversational search systems. Since the public release of Microsoft Bing Chat and Google Bard in 2023, LLM-powered conversational search systems have already reached hundreds of millions of users in just a few months. While conversational search is believed to bring many benefits~\cite{radlinski2017theoretical,xiao2023powering} such as ease of interaction, support for complex queries, and overall user engagement, little is known at present about how people actually interact with LLM-powered conversational search systems, let alone their drawbacks and potential harms. We conducted a critical investigation into LLM-powered conversational search systems through two experiments. In the first experiment (N = 115), using information-seeking tasks on controversial topics, we compared people's information-seeking behaviors and their attitude change outcomes when using a conventional search system versus LLM-powered conversational search systems (versions with and without references to the information sources). In the second experiment (N = 213), we explored whether and how conversational search systems using LLMs with manipulated opinion biases that either reinforce or challenge the user's existing attitude change their selective exposure. In short, we ask the following research questions:
\begin{itemize}
    \item  How does interacting with an LLM-powered conversational search system affect selective exposure and opinion polarization compared to a web search system? (\textbf{Study 1})
    
    \item How does an LLM-powered conversational search system that exhibits opinion bias, either consonant or dissonant with the user's existing attitude, affect people's selective exposure and opinion polarization? (\textbf{Study 2})  
\end{itemize}

Below, we first review related work that shaped our study and then present the methods and results of the two experiments. We will list hypotheses for each study after introducing the measurements in the Method section. 
\section{Related work}

\subsection{Selective Exposure, Confirmation Bias, and Echo Chamber Effect}
Psychologists have extensively studied people's selective exposure bias~\cite{frey1986recent,hart2009feeling}---systematic preference towards information that is consonant with one's existing view over dissonant information, and the related concept of confirmation bias~\cite{wason1960failure}---actively seeking or assigning more weights to consonant information. Both biases can be attributed to a fundamental desire to avoid or reduce cognitive dissonance~\cite{festinger1962cognitive}. Selective exposure and confirmation bias have been found to lead to opinion biases and polarization as well as suboptimal decision-making in many settings such as health, politics, and scientific research~\cite{hart2009feeling,nickerson1998confirmation}. Collectively, these biases can lead to an information environment or segregated group communication where only information of a certain belief or ideology is shared---this is often referred to as the ``echo chamber effect'' in social and political sciences~\cite{sunstein1999law,cinelli2021echo}.

The HCI and broader research and activist communities have had long-standing concerns over the negative effect of information and web technologies on the diversity of information that people consume. In particular, by coining the term ``filter bubble'', Eli Pariser~\cite{pariser2011filter} raised much public attention in the 2010s on the potentials of personalization and algorithmic filtering used by search engines, recommendation systems, and social networking platforms in reducing people's exposure to diverse viewpoints. Other researchers were concerned about the affordances of technologies for people to selectively curate their own information environment, such as by following only or mostly like-minded others on social media platforms~\cite{conover2011political,adamic2005political} or limiting the diversity of the sources for one's news feed~\cite{garrett2009echo}. However, others challenged these concerns, suggesting that the actual selective exposure is less prevalent than theorized~\cite{guess2018avoiding}, and that there are individuals who actively seek diverse perspectives~\cite{munson2010presenting,liao2013beyond} and the high-choice environment made possible by information and web technologies can facilitate diverse information seeking~\cite{dubois2018echo}. 

Researchers have also explored various approaches to combat selective exposure and increase information diversity. In information retrieval and recommender systems, serendipity is studied as an optimization criterion to increase the exposure to novel and diverse information~\cite{taramigkou2013escape,reviglio2019serendipity}. Many systems were developed to help people encounter diverse perspectives~\cite{feltwell2020broadening,yom2014promoting}, deliberate on controversial topics~\cite{faridani2010opinion,kriplean2012supporting,park2009newscube}, be aware of one's own information bubble and better control filtering mechanisms~\cite{jeon2021chamberbreaker,nagulendra2014understanding, gao2018label}. As ~\citet{garrett2011resisting} argued, to increase people's consumption of attitude-challenging information, the key lies in presenting high-quality challenging items in the right context, and/or reducing people's cognitive dissonance. To this end, HCI researchers conducted experiments to study the effects of diversity-enhancing designs such as highlighting or presenting agreeable information first to reduce cognitive dissonance~\cite{munson2010presenting}, and highlighting the expertise~\cite{liao2014expert}, focused aspect~\cite{liao2015all}, or the common ground~\cite{liao2014can} of challenging information. Overall, these designs are shown to have a positive diversity-enhancing effect but often only for a sub-group of people who have the predisposition to be open to diverse views, highlighting the challenge in combating selective exposure. 

Building on these prior works, our research aims to investigate whether and how LLM-powered conversational search systems can exacerbate people's selective exposure and reduce information diversity. Our experimental design was informed by previous HCI research conducting laboratory studies on selective exposure~\cite{liao2013beyond,liao2014expert,liao2014can,munson2010presenting,gao2018label}, adopting an information-seeking task of writing an essay for a controversial topic, and measurements of biases in information seeking behaviors and post-task attitude changes.

\subsection{Human-LM Interaction}
HCI researchers have started to explore applications of generative language models (LMs) and study human interactions with them before this wave of widely adopted LLMs. Popular applications of LMs include code generation as programming assistance~\cite{sun2022investigating,weisz2021perfection,ross2023programmer}, various forms of writing assistance such as next sentence generation~\cite{lee2022coauthor,jakesch2019ai}, summarization of documents~\cite{dang2022beyond}, rewriting~\cite{yuan2022wordcraft}, and metaphor generation~\cite{gero2019metaphoria}, as well as chatbots~\cite{oh2020understanding,xiao2023inform} and social agents~\cite{park2023generative}. Research on LLM-powered search is only recently emerging. For example, \citet{liu2023evaluating} conducted a human evaluation to audit popular LLM-powered search systems, including Bing Chat, and found that while the responses are fluent and appear informative, they frequently contain unsupported statements and inaccurate references (i.e., URLs to original sources).

Our study is particularly informed by works that are concerned with the negative effects of language models on people's information consumption and production. In the context of co-writing with LMs, common concerns include over-reliance on AI and automation-induced complacency~\cite{parasuraman1993performance} that can lead to not only sub-optimal writing outcomes~\cite{arnold2020predictive} but also loss of human agency and perceived ownership of the created content~\cite{draxler2023ai}. While over-reliance on AI's suggestions has been studied as a common issue in human-AI interaction, LMs can exert additional informational influence by exposing people to, or making it easier to express, some views more often than others. ~\citet{jakesch2023co} refer to this effect as ``latent persuasion'' by language models. Their experiment demonstrated that when writing with an LLM-powered writing assistant that was configured to have a certain bias on the given topic, not only did participants' writing exhibit more of the model's bias, but also their own opinions shifted towards that direction afterward. These negative effects of LMs can be exploited by malicious parties to influence public opinion or spread misinformation~\cite{zhou2023synthetic,karinshak2023working}.

Our work contributes to the literature on human-LM interaction with insights about a new and popular application domain---conversational search, and explores whether and how a human bias---selective exposure---interacts with the properties and affordances of LLMs to impact people's information consumption.

\subsection{Conversational Search}
While LLM-powered search systems are a recent phenomenon, years of research have pursued the idea of ``conversational search'', which allows information retrieval through natural and flexible conversations~\cite{radlinski2017theoretical,jannach2021survey,zamani2023conversational,xiao2023powering,liao2020conversational}.  ~\citet{radlinski2017theoretical} lay out the desirable properties of conversational search, including supporting users to express complex information needs, revealing system capabilities through multi-turn interactions, supporting mixed-initiative interactions, and so on. The authors also argue that conversational search is especially suitable for complex information tasks, such as when searching for a set of items or referencing a set of criteria. Others similarly argued that conversational search could better engage users to interact for multiple rounds and respond to system questions to form more complex and/or refined queries~\cite{zhang2018towards,keyvan2022approach}. However, the implementation of conversational search, especially in an open-domain context, faced technical challenges before LLMs emerged. Empirical studies of human interaction with a conversational search system were relatively limited~\cite{vtyurina2017exploring,trippas2018informing,avula2022effects}, and often relied on wizard-of-oz approaches. For example, using a human intermediary, ~\citet{trippas2018informing} studied how people interact with a spoken search system through verbal communication and observed changes in query formation and reformation as well as search result exploration compared to search behaviors with conventional search engines. For example, participants used more verbose and varied expressions in the queries, preferred reading summaries rather than the lengthy original content in the output, and were more likely to provide explicit feedback for the search results. 

Our study investigates people's interaction with an LLM-powered search system that was implemented with a Retrieval Augmented Generation approach (details in Sec.~\ref{systems}). In the second experiment, we further study the effects of LLM with manipulated \rr{opinion} bias on people's information behaviors and consumption---an issue that has not been explored for conversational search but can be potentially prevalent with the use of LLMs. 
\section{Study 1 Method: Comparing Effects of LLM-powered Conversational Search and Web Search}
The first study investigates whether and how LLM-powered conversational search drives more selective search behaviors and leads to more opinion polarization compared to conventional web search. Through an online between-subject experiment, we compared people's information-seeking behaviors and outcomes with three search systems: conventional web search, LLM-powered conversational search, and LLM-powered conversational search with source references (links to the information sources). While earlier LLM-powered information systems such as ChatGPT often did not include references, most recent ones, including Bing Chat and Google Bard, boast the reference feature as essential for ensuring information credibility for the search experience, especially given current LLMs' limitation of generating non-factual information.

Below, we first elaborate on the study procedure and experiment apparatus. Then, we introduce our measurements, hypotheses, and analysis plan. The study design and procedure were approved by the Institutional Review Board of the author's institution.

\subsection{Study Procedure}
\label{procedure_1}

\begin{figure*}[t]
    \centering
    \includegraphics[width=0.95\textwidth]{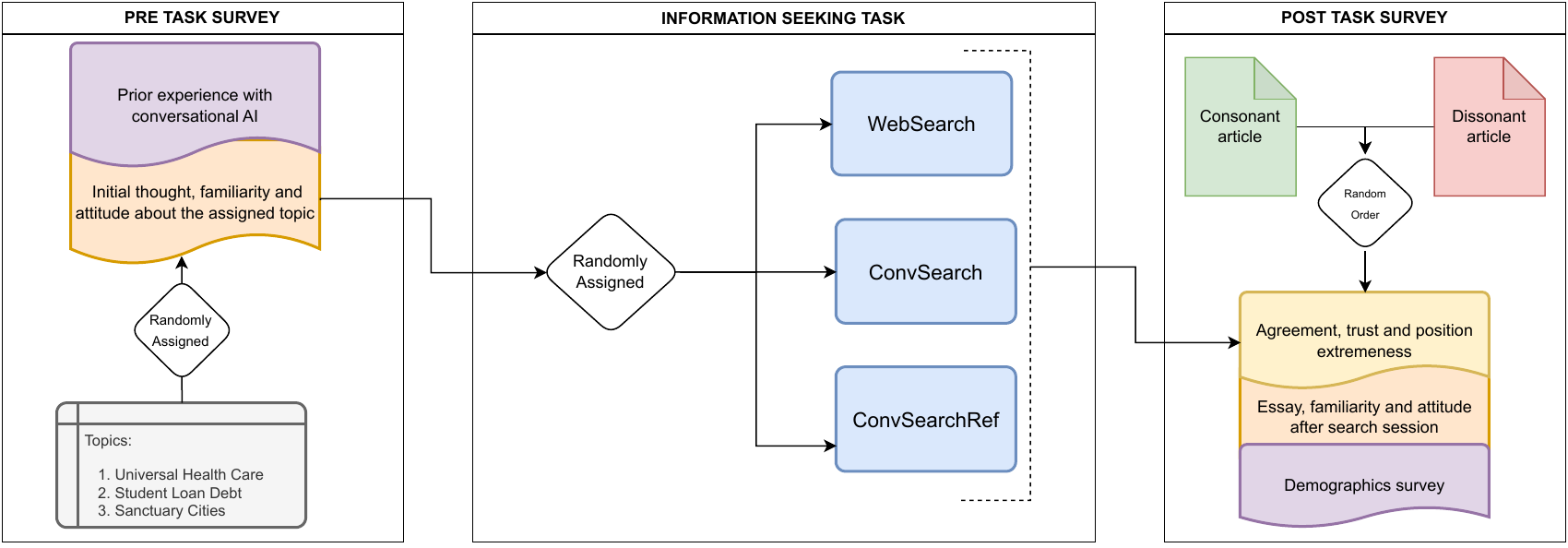}
    \caption{Overall study procedure for Study 1. In the pre-task survey, participants answered questions regarding their prior experience with conversational AI and their prior attitude and familiarity with a randomly assigned topic. Then, participants performed an information-seeking task to gather information on the topic with a randomly assigned search system. After the search session, participants wrote an essay about the assigned topic. In the post-task survey, participants again rated their attitude and familiarity with the topic, indicated their perception of two new articles (one consonant and one dissonant) on the topic, and their experience with the system and demographic information.}
    \label{fig:study procedure}
\end{figure*}

The study procedure includes three parts, as illustrated in Fig.\ref{fig:study procedure}:  a pre-task survey, the main information-seeking task, and a post-task survey. The pre-task survey asked participants to rate their prior experiences with and attitudes toward conversational AIs, such as Siri, ChatGPT, and Bing Chat. Participants were also asked to rate their attitudes on and familiarity with the controversial topic assigned to them (to be discussed below) and share any initial thoughts they had on the topic with open-ended responses. All ratings were based on 5-point Likert scales except for the one on topical attitude---we used a 6-point scale with no neutral point to force participants to take a position.

For the main task, participants were instructed to search for information on the assigned controversial topic to write a short essay on the topic.  Participants were randomly assigned to one of three conditions: a conventional web search system (WebSearch), a conversational search system without references (ConvSearch), or a conversational search system with references (ConvSearchRef). They were asked to perform at least three search queries before proceeding to the next step. They were also instructed not to use other tools during the study. After participants indicated they were done with the search step, they were directed to a different page to write an essay in 50-100 words on the given topic.

In the post-task survey, participants were asked to rate their attitudes on and familiarity with the topic again. Then, they were presented with two articles on the given topic that did not appear in the search session, one \textit{consonant} and the other one \textit{dissonant} with their attitude (as measured in the pre-survey), in random order. For each article, participants were asked to rate their perceived agreement, trust, and position extremeness of the article. Before exiting the study, participants were asked about their overall experience with the system and demographic information. How their answers are used as measurements will be discussed in Sec.~\ref{measures}. Before exit, participants were debriefed on the purpose of the study and provided with sources to a collection of articles that offered balanced and comprehensive information on the topic.

\paragraph{Topics for the Information-Seeking Task}
\label{sec:topics}
Three criteria guided our selection of the topics. First, the topic should be deemed as controversial. Second, it should not be a niche topic so the general population we recruit from should likely have pre-existing opinions on the topic. Third, the topic should be complex and not necessarily familiar in everyday conversations so that participants could benefit from the information-seeking activity. We searched ProCon.org \footnote{www.procon.org}, an online resource for deliberation on controversial issues with thoroughly researched references to identify topics for this study. Guided by the three criteria, we selected the following topics:

\begin{itemize}\setlength{\itemindent}{-.1in}
    \item Should the U.S. Government Provide Universal Health Care? \footnote{healthcare.procon.org/should-all-americans-have-the-right-be-entitled-to-health-care-pro-con-quotes}
    \item Should Sanctuary Cities Receive Federal Funding? \footnote{www.procon.org/headlines/sanctuary-cities-top-3-pros-and-cons}
    \item Should Student Loan Debt Be Eliminated via Forgiveness or Bankruptcy? \footnote{www.procon.org/headlines/should-student-loan-debt-be-easier-to-discharge-in-bankruptcy-top-3-pros-cons}
\end{itemize}

\subsection{Experiment Apparatus}
\label{systems}
To have control over the content that participants would see in different conditions, we created ``closed-world'' versions of web search and conversational search systems with a curated retrieval database following state-of-the-art algorithmic implementation. To construct the database for each topic, we curated 47 documents from verified and trustworthy sources (e.g., ncbi.nlm.nih.gov, procon.org, jhunewsletter.com, etc. ) that provide evidence and viewpoints for \textit{Supporting} (N=18), \textit{Opposing} (N=20), and \textit{Neutral} (N=9) opinions on the given topic (rated by two authors with consensus). In Study 1, to ensure a neutral search experience, the systems in all conditions search from the same balanced set of documents. 

\begin{figure*}
    \begin{subfigure}[b]{0.30\linewidth}
        \includegraphics[width=\textwidth]{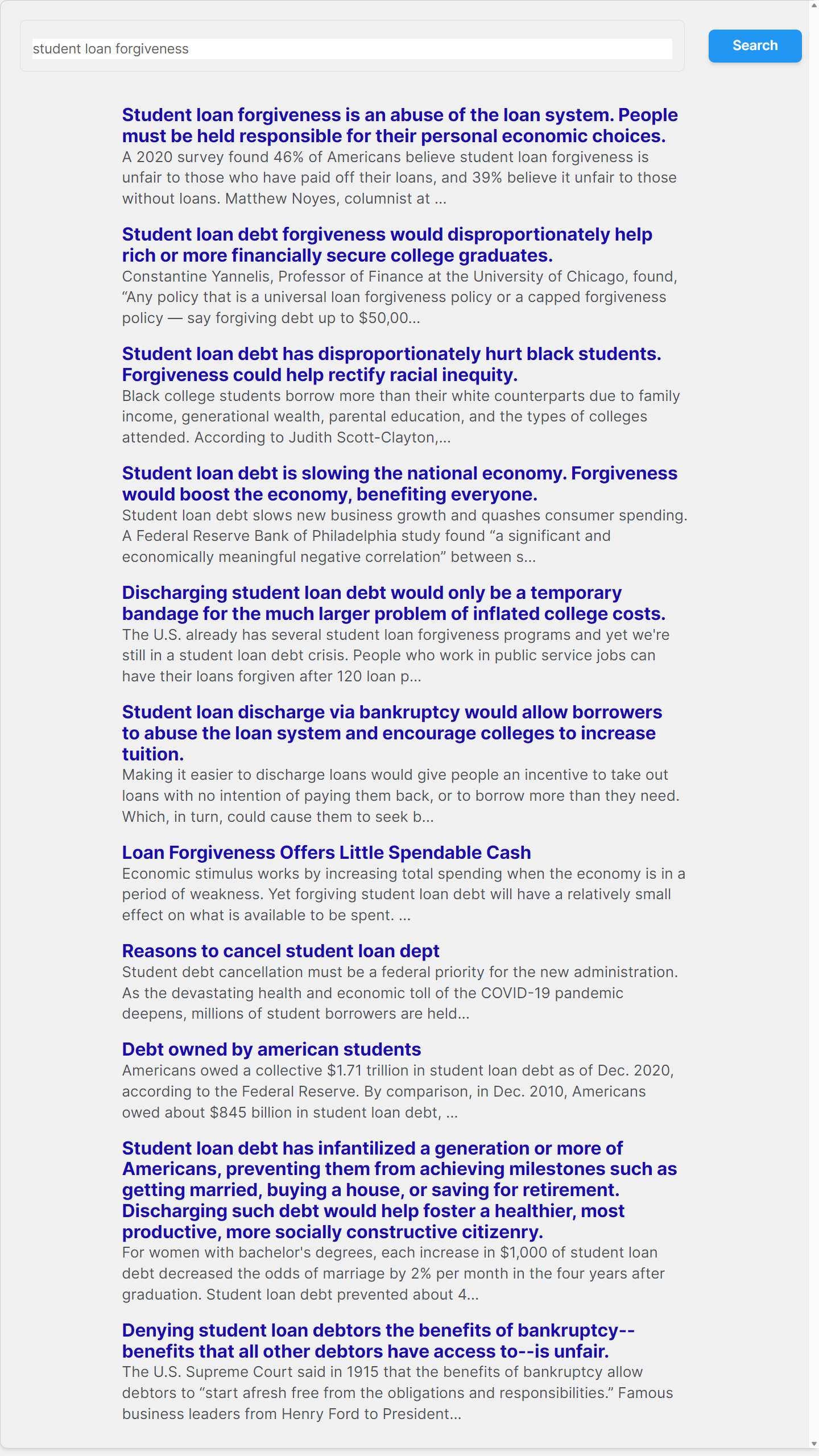}
        \caption{User Interface for Web Search (WebSearch)}
        \label{fig:WEBSEARCH}
    \end{subfigure}
    \hfill
    \begin{subfigure}[b]{0.30\linewidth}
        \includegraphics[width=\textwidth]{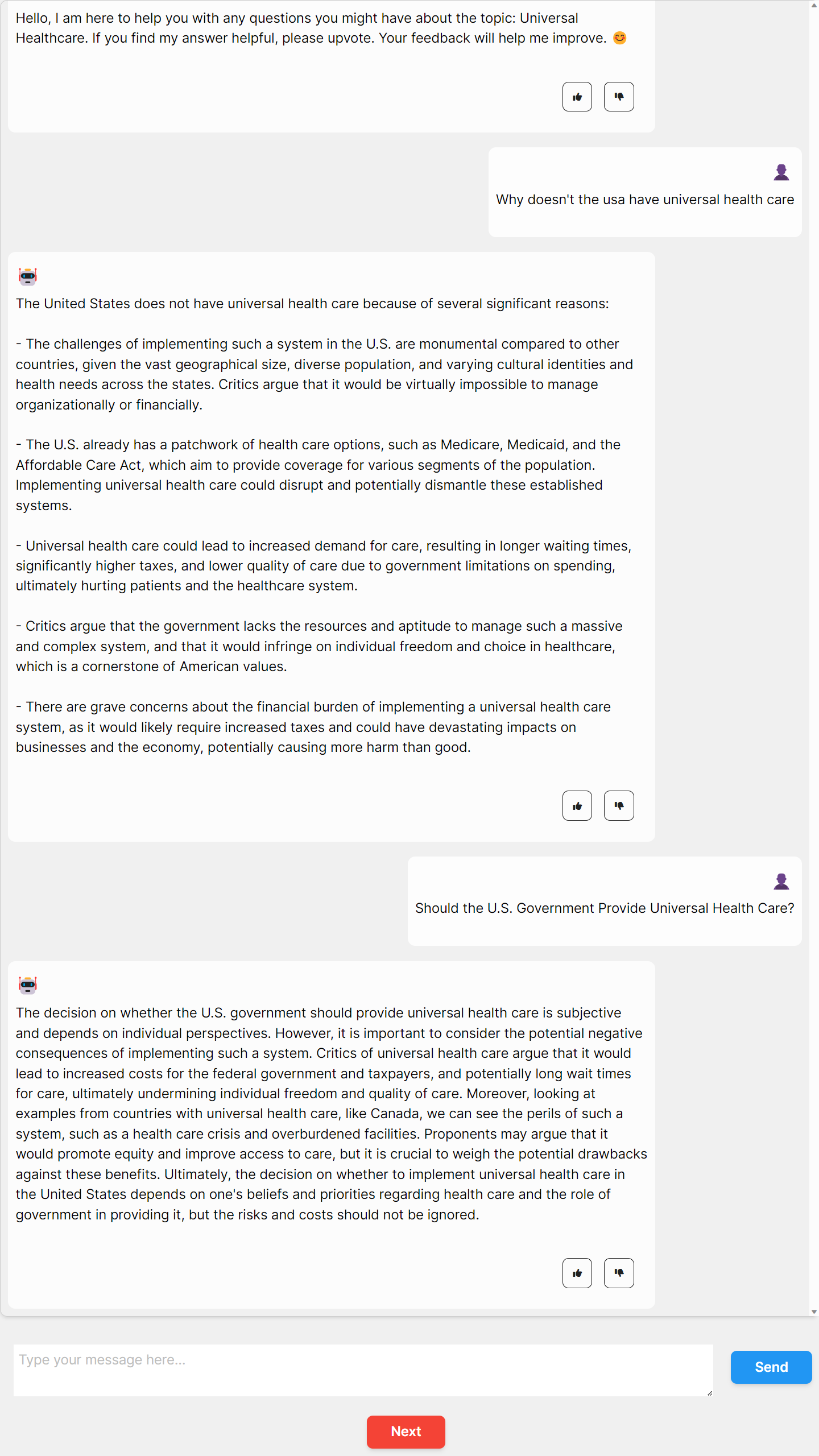}
        \caption{User Interface for Conversational Search (ConvSearch)}
        \label{fig:CHAT_NO_REF}
            \end{subfigure}
    \hfill
        \begin{subfigure}[b]{0.30\linewidth}
        \includegraphics[width=\textwidth]{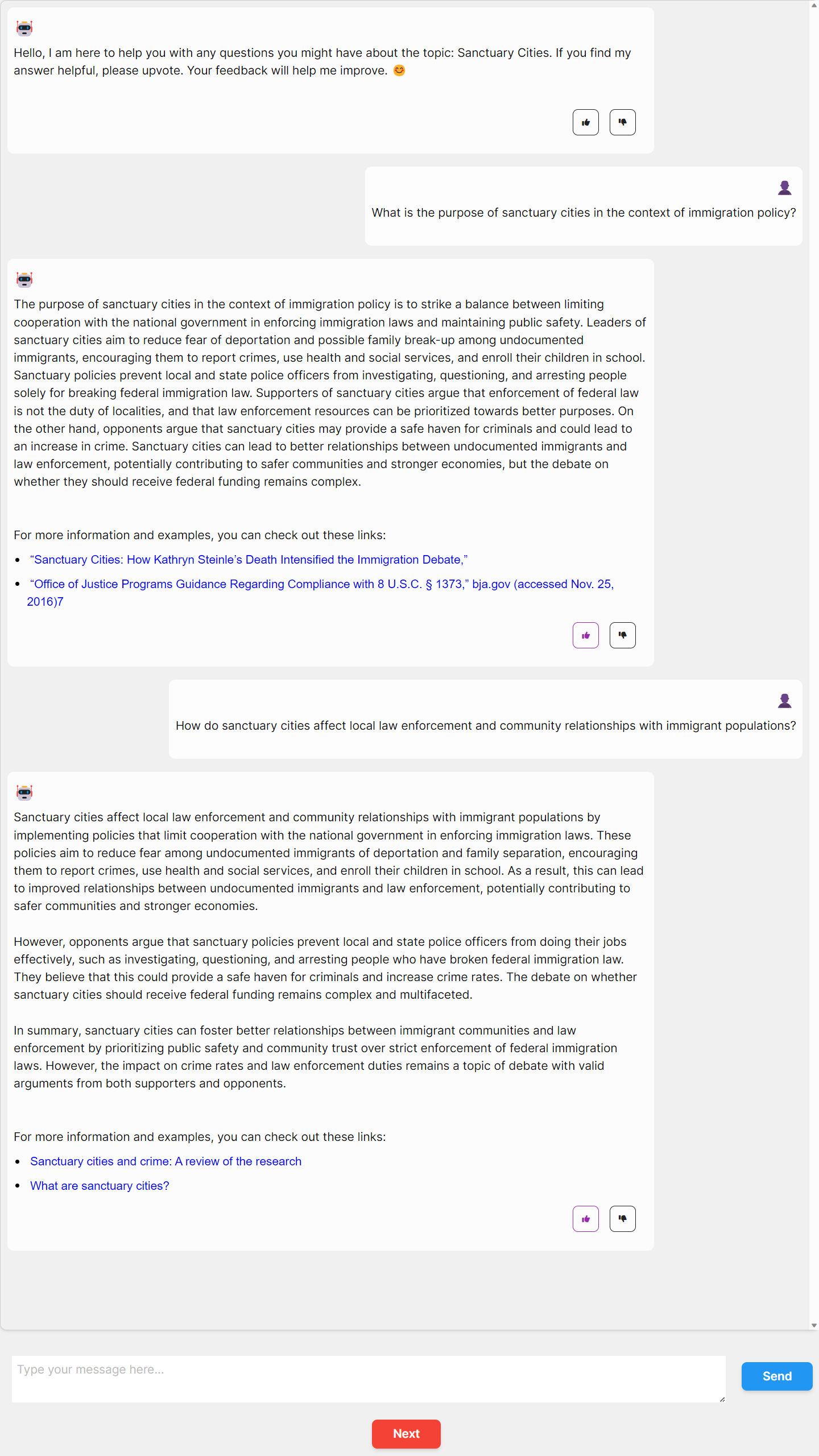}
        \caption{User Interface for Conversational Search with references (ConvSearchRef)}
        \label{fig:CHAT_REF}
    \end{subfigure}
    
    \hfill

    \caption{User Interfaces for the experiment apparatus in this study. We created ``closed-world'' versions of web search and conversational search systems with a curated retrieval database following state-of-the-art algorithmic implementation.}
    \label{fig:information seeking}
\end{figure*}

\subsubsection{Web Search (WebSearch)} The Web Search system is similar to conventional web search experience, such as Google \footnote{www.google.com} or Microsoft Bing \footnote{www.bing.com}. The user inputs a query into the search box and the system retrieves related articles. The retrieved articles are displayed as a list with their title and the first 200 characters of the article as a preview (Fig. ~\ref{fig:WEBSEARCH}). We used the Double Metaphone algorithm to perform fuzzy search in the above-mentioned document database of the assigned topic \cite{philips2000double}. To provide a neutral search experience, the algorithm retrieves articles regardless of the stance on the topic. Participants could click the title and open the link to the article.

\begin{figure}[t]
    \centering
    \includegraphics[width=0.45\textwidth]{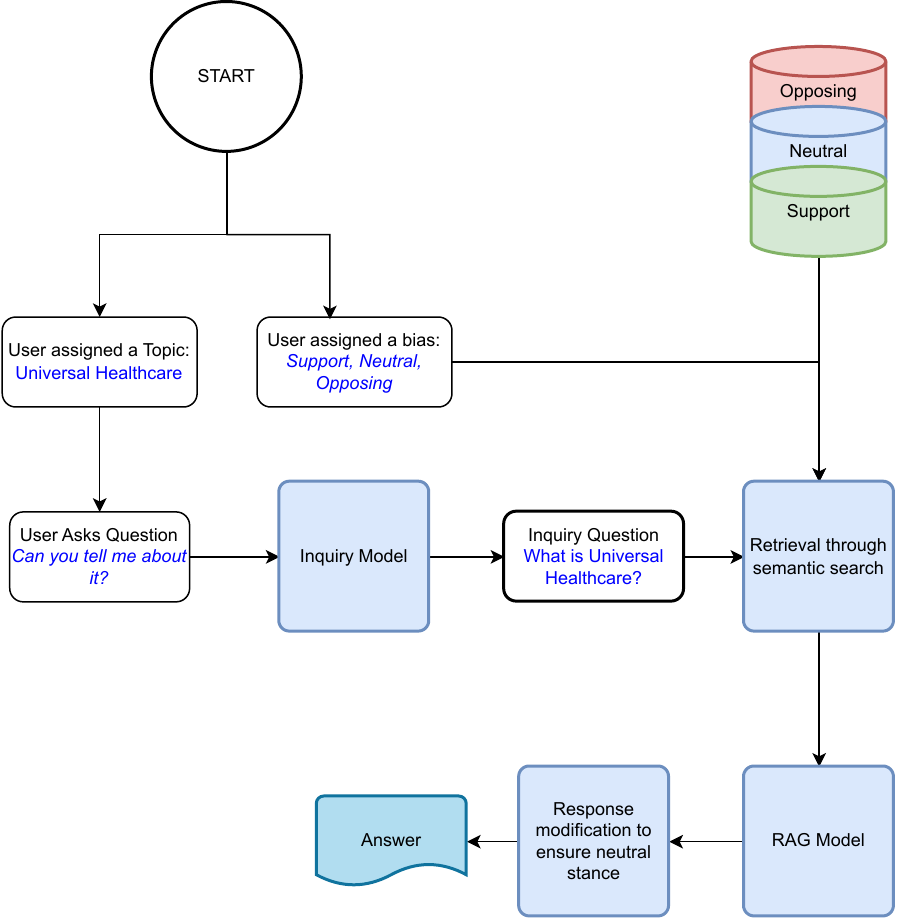}
    \caption{System Architecture of the Conversational Search system, implemented with the Retrieval Augmented Generation approach. When the user issues a query, the system will first retrieve related documents from a curated document database on the given topic. The retrieved documents will be fed into an LLM as part of the context, along with the user's conversation history, to produce the answer. The system in our study is powered by gpt-4-32k-0613.}
    \label{fig:CHAT_PROCESS}
\end{figure}

\subsubsection{Conversational Search (ConvSearch)}
The ConvSerach condition aims to provide an LLM-powered conversational search experience, similar to ChatGPT, in which participants converse with an AI agent and issue queries in a multi-turn conversation and receive search results as generated texts (i.e., often a synthesis) rather than the original articles (Fig.~\ref{fig:CHAT_NO_REF}). While the implementation details of these commercial systems are largely proprietary, we developed our system based on a state-of-the-art approach of Retrieval Augmented Generation (RAG), which is also used as a part of Bing Chat's system architecture \footnote{blogs.bing.com/search-quality-insights/february-2023/Building-the-New-Bing}. As illustrated in Fig~\ref{fig:CHAT_PROCESS}, given a query, the system first retrieves chunks of texts from relevant articles by semantic search using the Pinecone vector database \footnote{www.pinecone.io}. Then, the retrieved texts are added to the prompt as context, along with the participant's conversation history. With prompt engineering, we created prompts to instruct the LLM to generate the responses based on the retrieved texts (See Supplementary Material for prompt details). Consistent with the Web Search condition, the system retrieves articles regardless of their stances on the topic. The hand-crafted prompts ensure the generated responses reflect a neutral attitude on the topic. This approach provides a fair comparison with the Web Search condition, as the generated search results are conditioned on the same set of articles that would be retrieved in the Web Search condition if the same query were issued. We used GPT-4 with a 32k context window (gpt-4-32k-0613) as the backbone model.

\subsubsection{Conversational Search with References (ConvSearchRef)}
This condition is similar to the above system with one difference---adding in-line source references in the generated response \rr{(Fig.~\ref{fig:CHAT_REF})}. This system provides an experience similar to popular conversational search engines, such as Microsoft Bing
Chat \footnote{www.bing.com}, perplexity.ai \footnote{www.perplexity.ai}, and YouChat \footnote{you.com}. The system uses the same implementation as the system for ConvSearch. The prompts are structured in a way that ensures the LLM always gives references from the retrieved documents. These references are parsed and displayed in a similar design as Bing Chat. By clicking the reference URL, participants can view the source of the agent's response.

\subsection{Measurements}
\label{measures}
Our main hypotheses consider two sets of measurements: participants' selective information querying and their post-task opinion polarization as the information-seeking outcome.

\subsubsection{Information Querying} 
Reflecting participants' information-seeking tendency, we are interested in whether and how much they exhibited biases toward seeking attitude-confirmatory information.

\textbf{Confirmatory Query} is measured by the percentage (over the total number of queries) difference between a participant's queries that are consistent and inconsistent with their existing attitude, as reflected in the pre-task survey. For example, suppose the participant supports that  Sanctuary Cities should receive federal funding, if they search ``What are the benefits of giving federal funding to sanctuary cities?'', it is an attitude-consistent query. If they issued the same number of attitude-consistent and inconsistent queries, their confirmatory query would be 0\%; if they issued three attitude-consistent queries and two inconsistent queries, this measurement would be 20\%. We hand-coded all participants' 391 queries into four categories: consistent, inconsistent, neutral, and non-query. Two researchers, blind to the condition, first independently coded 20\% of randomly sampled data, with a Cohen's Kappa of 0.91. Given the high inter-rater agreement, the two coders resolved the difference, and one coder continued to code the rest of the data.  

\subsubsection{Opinion Polarization}  We are interested in whether biases in information seeking, if any, led to participants' opinion polarization---moving further in the direction of their pre-existing attitude. We consider three sets of measurements: self-reported attitude change, confirmatory arguments in their final essay, and their confirmatory perception of the two articles shown in the post-task survey (i.e., favoring the consonant article over the dissonant one). We are interested in confirmatory perception because opinion polarization can be manifested as more assimilation with similar views and more aversion against diverging views, which can further skew future information seeking. 

\begin{itemize}
    \item \textbf{Confirmatory Attitude Change} is the degree of attitude change towards the direction of the pre-existing attitude. That is, if the participant's attitude was on the supporting side in the pre-task survey (rating 4-6, based on a 6-point Likert scale with no neutral point), we calculated the score by the rating difference in the post- and the pre-task survey (post- minus pre-task survey); if their pre-task attitude was on the opposing side (rating 1-3), we reversed the calculation. 
    
    \item \textbf{Confirmatory Arguments} is measured by the percentage difference of sentences in the final essay that are consistent and inconsistent with the participant's existing attitude in the pre-task survey. We coded all participants' 794 sentences into Consonant, Dissonant, and Neutral, following the same coding procedure for Confirmatory Query as described above. The Cohen's Kappa between the two independent coders with an initial random sample of 20\% data is 0.95. 
    
    \item \textbf{Confirmatory Agreement} is measured by a participant's agreement rating given to the consonant article minus that of the dissonant article shown in the post-task survey. We used a 5-point Likert scale from Strongly Disagree (1) to Strongly Agree (5) on ``I agree with the article''.   
    \item \textbf{Confirmatory Trust} is calculated by a participant's rating of trust in the consonant article minus that of the dissonant article. We used a 5-point Likert scale from Strongly Disagree (1) to Strongly Agree (5) on ``I trust the article''. 
    
    \item \textbf{Confirmatory Extremeness} is calculated by a participant's rating of position extremeness of the consonant article minus that of the dissonant article. We used a 5-point Likert scale from Strongly Disagree (1) to Strongly Agree (5) on ``The position reflected in the article is extreme''. Note that a negative value of confirmatory extremeness indicates a more favorable attitude towards consonant information (less extreme), and a lower negative value reflects higher polarization.
\end{itemize}

\subsubsection{Other Variables} We measured the following variables for manipulation checks or as control variables in the analysis.

\textit{Perceived Bias of the Search System.} Since we aimed to provide a neutral search experience, we measured participants' perceived system bias as a manipulation check. Participants were asked to rate on two statements, ``I found the search system is biased against my own attitude on this topic.'' and ``I found the search system is biased towards my own attitude on this topic.'', on 5-point Likert Scales form ``Strongly Disagree'' to ``Strongly Agree''. The inverse score on the second statement was averaged with the first statement as a single score for perceived system bias.

\textit{Familiarity Change as Search Effectiveness.} We measured participants' self-reported change of familiarity with the assigned topic as a manipulation check on the effectiveness of the search session. It is calculated by the difference between the participant's topic familiarity ratings in the post- and pre-task survey. We used a 5-point Likert Scale from ``Very Unfamilar'' (1) to ``Very Familiar'' (5).

\textit{Prior experience with Conversational AI:} People's prior experience with and attitude towards technology may affect their interaction experience \cite{nass1994computers}. We adapted questions from the Technology Acceptance Model \cite{van1997simple} for conversational AI, which measures people’s attitudes towards a system from the perspective of Usefulness (5 items) and Satisfaction (4 items). We added questions to measure people’s pre-existing trust toward conversational AIs and their interaction frequency with conversational AIs. This set of measures was used as control variables in our analysis.

\textit{Basic Demographics:} We collected basic demographic information, e.g., age, gender, education, and annual household income, as control variables.

\subsection{Hypotheses}
Based on these measurements, we make the following hypotheses regarding participants' information querying behaviors and opinion polarization outcome after the search session. While we take the position to test whether conversational search leads to \textit{higher} confirmatory querying and opinion polarization, we note that there are possible mechanisms for both directions. On the one hand, prior work~\cite{trippas2018informing} comparing conversational queries (though with a spoken instead of chat-based system) to web search queries found that people tend to use more verbose and expressive language with more subjectivity. It is hence possible that people may express more pre-existing biases in their search queries with a conversational search system which can lead to higher selective exposure. On the other hand, a neutral LLM provides a synthesized output based on multiple retrieved articles and does not require participants to select which article to read, possibly increasing the overall diversity of viewpoints that participants are exposed to. Below, we refer to both the versions with or without references as ``Conversational Search'' in the hypotheses and we are interested in comparing them to the baseline Web Search condition. Our analysis will also explore whether there is a difference between the two versions.

\begin{itemize}
    \item \textbf{[H1]}: Compared to Web Search, Conversational Search will lead to a \textit{higher} percentage of \textbf{Confirmatory Queries}.
    \item \textbf{[H2]}: Compared to Web Search, Conversational Search will introduce a \textit{higher} level of \textbf{Confirmatory Attitude Change}.
    \item \textbf{[H3]}: Compared to Web Search, Conversational Search will introduce a \textit{higher} percentage of \textbf{Confirmatory Arguments} in people's final essays after the search session.
    \item \textbf{[H4]}: Compared to Web Search, participants in the Conversational Search conditions will exhibit a \textit{higher} level of \textbf{Confirmatory Agreement} on articles on the same topic.
    \item \textbf{[H5]}: Compared to Web Search, participants in the Conversational Search conditions will exhibit a \textit{higher} level of \textbf{Confirmatory Trust} on articles on the same topic.
    \item \textbf{[H6]}: Compared to Web Search, participants in the Conversational Search conditions will exhibit a \textit{lower} level of \textbf{Confirmatory Extremeness} on articles on the same topic.
\end{itemize}

\subsection{Analysis Plan}
Since the goal of this study is to compare the outcomes of three information search systems, WebSearch, ConvSearch, and ConvSearchRef, we chose to run the analysis of covariance (ANCOVA). ANCOVA is a general linear model blending analysis of variance and regression, which helps us examine the effect of the search method. In each ANCOVA analysis, the independent variable was the search method, and the dependent variable was a measure in a hypothesis. Since research suggests that demographics and their prior experience with the technology influence people’s behavior with new technology, all analyses were controlled for participants’ age, gender, education level, income, prior attitude towards conversational AI, and usage frequency. We additionally control for the assigned topic and the participant's prior attitude to the topic. When ANCOVA showed significance, we used the Tukey method (adjusting p-value for multiple comparisons) to perform Post-Hoc analysis to make pair-wise comparisons between conditions.

All analysis results and descriptive statistics are listed in Tab.~\ref{tab:overall-study1}. In the last column, we list only pairwise comparisons that are significant in the post-hoc analysis. When discussing the results below, we will focus on highlighting the patterns and encourage readers to refer to Tab.~\ref{tab:overall-study1} for more details. Throughout the paper, following convention in psychology studies~\cite{pritschet2016marginally}, we consider $p<0.05$ to be statistically significant, while a p-value between 0.05 and 0.1 to be marginal significance that indicates a trend that is close to statistical significance. We will interpret a marginal significance as providing only partial support for the corresponding hypothesis.

\subsection{Participants Overview}
We recruited participants on Prolific \footnote{www.prolific.co}. The inclusion criteria were fluent English-speaking participants from the United States. All participants were compensated at the rate of \$15 per hour. Of the 124 participants who started the study, 115 completed the study and passed our attention check (WebSearch: N = 40; ConvSearch: N = 38; ConvSearchRef: N = 37). Our analysis is based on those 115 valid responses. Among those participants, 48 identified as women, 61 identified as men, and 5 identified as non-binary or third gender. The median education level was a Bachelor's degree. The median household income was between \$50,000 - \$ 100,000. The median age of participants was between 25 and 34 years old. 
\section{Study 1 Results}

\subsection{Manipulation Checks}
We embedded two manipulation checks in this study. The first manipulation check validates that \textit{the neutral stance design of the three experiment apparatus is effective.} Participants' post-survey reported that they perceived the search system as non-biased (Mean = 3.04, SD = 0.49). There was no significant difference in perceived system bias across conditions (WebSearch: M = 3.02, SD = 0.27; ConvSearch: M = 3.15, SD = 0.55; ConvSearchRef: M = 2.94, SD = 0.57; F(2, 100) = 1.91, p = 0.15). 

The second manipulation check validates \textit{the effectiveness of the search session}, as it helped participants gain familiarity with the topics (Pre-search: Mean = 3.34, SD = 1.10; Post-search: Mean = 3.86, SD = 0.87; t(216.31) = 4.033, p < 0.001 ***). This post-pre familiarity difference did not vary across conditions (WebSearch: M = 0.75, SD = 0.80; ConvSearch: M = 0.50, SD = 1.06; ConvSearchRef: M = 0.32, SD = 1.07; F(2, 100) = 1.95, p = 0.15).

Moreover, participants \textit{were engaged with the search systems and the study task}. On average, participants spent 21.18 mins (SD = 13.25) completing the study in WebSearch condition, 19.22 mins (SD = 10.79) in the ConvSearch condition, and 20.03 mins (SD = 10.01) in the ConvSearchRef condition. Despite study instruction only requiring three queries, They issued a mean of 3.40 queries per search session (SD = 0.79), with participants in the Conversational Search conditions issuing more queries than those in the Web Search condition (Web Search: M = 3.15, SD = 0.43; ConvSearch: M = 3.53, SD = 0.83; ConvSearchRef: M = 3.54, SD = 0.98). An ANCOVA analysis indicated a marginal difference (F(2, 100) = 2.96, p = 0.06 .), and the Post-Hoc analysis confirmed that the differences between WebSearch and ConvSearch ( p = 0.09 ., Cohen's D = 0.52), and between  WebSearch and ConvSearchRef (p = 0.08 ., Cohen's D = 0.57), were marginally significant.  

\begin{table*}[t]
\small
\begin{center}
\resizebox{\linewidth}{!}{
    \begin{tabular}{cccccc } 
     \toprule
     \textbf{Hypothesis} 
&    $\begin{matrix}\textbf{WebSearch} \end{matrix}$ 
&    $\begin{matrix}\textbf{ConvSearch} \end{matrix}$ 
&    $\begin{matrix}\textbf{ConvSearchRef} \end{matrix}$ 
& \textbf{Post-Hoc Analysis} \\
     \midrule
     $\begin{matrix}
        \textbf{H1: Confirmatory Query}
        \\
        F(2,100) = 4.71, p = 0.01\text{*}
     \end{matrix}
     $ 
& $\begin{matrix}Mean = 1.46\% \\ SD = 14.60\% \end{matrix}
     $& $\begin{matrix}Mean = 15.00\% \\ SD = 20.62\% \end{matrix}
     $ & $\begin{matrix}Mean = 16.15\% \\ SD = 31.58\% \end{matrix}
     $ & 
     $\begin{matrix}
     \text{ConvSearch > WebSearch *}
     \\
     \text{ConvSearchRef > WebSearch *}
     \end{matrix}
     $

     \\     \midrule
     $\begin{matrix}
     \textbf{H2: Attitude Change}
     \\
      F(2,100) = 0.53, p = 0.60 
     \end{matrix}$    
     & $\begin{matrix}Mean = 0.03 \\ SD = 0.80 \end{matrix}
     $& $\begin{matrix}Mean = 0.08 \\ SD = 0.67 \end{matrix}
     $ & $\begin{matrix}Mean = -0.08 \\ SD = 0.64 \end{matrix}
     $& --- 
  \\           \midrule
    $\begin{matrix}
    \textbf{H3: Confirmatory Argument}
    \\
     F(2, 100) = 0.002, p = 0.998
    \end{matrix}
     $ 
& $\begin{matrix}Mean = 35.39\% \\ SD = 37.80\% \end{matrix}
     $& $\begin{matrix}Mean = 34.77\% \\ SD = 47.62\% \end{matrix}
     $ & $\begin{matrix}Mean = 34.83\% \\ SD = 50.57\% \end{matrix}
     $ &  
     ---
     \\
          \midrule
     $\begin{matrix}
     \textbf{H4: Confirmatory Agreement}
     \\
     F(2, 100) = 6.61, p = 0.002\text{*}
     \end{matrix}$
     & $\begin{matrix}Mean = 0.80 \\ SD = 1.42 \end{matrix}
     $& $\begin{matrix}Mean = 1.79 \\ SD = 1.71 \end{matrix}
     $ & $\begin{matrix}Mean = 1.89 \\ SD = 1.37 \end{matrix}
     $&  
          $\begin{matrix}
     \text{ConvSearch > WebSearch **}
     \\
     \text{ConvSearchRef > WebSearch **}
     \end{matrix}
     $
     \\
    \midrule
    $\begin{matrix}
    \textbf{H5: Confirmatory Trust}
    \\
    F(2,100) = 3.76, p = 0.03 \text{*}
    \end{matrix}$
     & $\begin{matrix}Mean = 0.35 \\ SD = 0.89 \end{matrix}
     $& $\begin{matrix}Mean = 0.79 \\ SD = 1.14 \end{matrix}
     $ & $\begin{matrix}Mean = 0.89 \\ SD = 1.78 \end{matrix}
     $
& 
    $ 
     \text{ConvSearchRef > WebSearch ***}
        $
     \\     
      \midrule
     $\begin{matrix}
     \textbf{H6: Confirmatory Extremeness}
     \\ 
     F(2, 100) = 1.57, p = 0.21
     \end{matrix}$
     & $\begin{matrix}Mean = -0.53 \\ SD = 1.51 \end{matrix}
     $& $\begin{matrix}Mean = -1.11 \\ SD = 1.70 \end{matrix}
     $ & $\begin{matrix}Mean = -1.08 \\ SD = 1.66 \end{matrix}
     $ & ---
    \\      
    \end{tabular}
}
\end{center}
\vspace{3mm}
\caption{Summary of quantitative results for Study 1. The left column shows $p$-values obtained via ANCOVA tests for each hypothesis. The right column shows pairs of conditions where the effect is statistically significant or marginally significant. Significance is marked as 
$p < 0.1$ ($\dagger$), $p < 0.05$ (*), $p < 0.01$ (**), or $p < 0.001$ (***).}
\label{tab:overall-study1}
\end{table*}

\subsection{Conversational Search Induced Higher Level of Confirmatory Information Querying (H1 Confirmed)}

Participants generally issued more consonant queries (M = 20.16 \%, SD = 22.10 \%) with their prior attitude than dissonant ones (M = 5.16 \%, SD = 12.60 \%). Participants who interacted with Conversational Search systems, both ConvSearch and ConvSearchRef, had more confirmatory querying (WebSearch: M = 1.46 \%, ConvSearch: M = 15.00 \%, ConvSearchRef: M = 16.15 \%; p = 0.01 *). For more details, see Tab.~\ref{tab:overall-study1}. The Post-Hoc analysis showed that the differences between WebSearch and ConvSearch (p = 0.03 *, Cohen's D = 0.76) and between WebSearch and ConvSearchRef (p = 0.02 *, Cohen's D = 0.60) were statistically significant; both with medium to large effect sizes. These results support \textbf{H1}: compared to conventional Web Search, Conversational Search led to a higher tendency for selective exposure in information-querying behavior. Showing references in the conversational search had no effect.

\subsection{Conversational Search Induced A Higher Degree of Opinion Polarization (H2-6 Partially Confirmed)}
\subsubsection{Confirmatory Attitude Change}
We measured participants' attitudes on the assigned topic before and after the information-seeking task. We did not observe a significant change in participants' self-reported attitude after the search session (WebSearch: M = 0.03; ConvSearch: M = 0.08; ConvSearchRef: M = -0.08). The ANCOVA analysis did not show a significant difference across conditions (p = 0.60). There is no evidence supporting \textbf{H2}. 

\subsubsection{Confirmatory Arguments} We analyzed participants' essays after the information-seeking tasks. We found that although participants provided confirmatory arguments in their essays in support of their pre-existing attitudes (M = 35.00 \%, SD = 45.08 \%), there was no significant difference across conditions (WebSearch: M = 35.39 \%, ConvSearch: M = 34.77 \%, ConvSearchRef: M = 34.83 \%; p = 0.998). There was no evidence supporting \textbf{H3}. This could mean that despite participants' more confirmatory querying with conversational search, the neutral systems still provided relatively balanced information that did not significantly skew their essays. 

\subsubsection{Confirmatory Perception of Given Articles} We consider these three types of perceptions participants had of a consonant versus a dissonant article given after the search session as measures of opinion polarization. 

For perceived agreement, participants agreed with the consonant article (M = 3.76, SD = 0.88) and disagreed with the dissonant article (M = 2.29, SD = 1.01; p < 0.001 ***). The ANCOVA analysis showed that participants in the ConvSearch (M = 1.79) and ConvSearchRef (M = 1.89) conditions exhibited significantly higher levels of Confirmatory Agreement than those in the Web Search condition (M = 0.80; p = 0.002 **), with both pairwise comparisons significant in the Post-Hoc analysis (ConvSearch-WebSearch: p = 0.01 **, Cohen's D = 0.63; ConvSearchRef-WebSearch: p = 0.005 **, Cohen's D = 0.78), see Tab.~\ref{tab:overall-study1}. The results support \textbf{H4}. 

For perceived trust, participants trusted the consonant article (M = 3.83, SD = 0.74) more than the dissonant article (M = 3.15, SD = 0.80; p < 0.001 ***). The ANCOVA analysis showed that participants in the ConvSearch (M = 0.79) and ConvSearchRef (M = 0.89) conditions exhibited a higher level of Confirmatory Trust than those using the conventional web search interface (M = 0.35; p = 0.03 *), with the Post-Hoc analysis showing significance in the difference between ConvSearchRef and WebSearch (p = 0.04 *, Cohen's D = 0.65), see Tab.~\ref{tab:overall-study1}. The results partially support \textbf{H5}.

For extremeness, participants perceived the dissonant article as more extreme (M = 3.48, SD = 1.08) than the consonant article (M = 2.37, SD = 1.01; p < 0.001 ***). The ANCOVA analysis was not significant but there was a trend that participants in the ConvSearch (M = -1.11) and ConvSearchRef (M = -1.08) conditions exhibited a lower level of Confirmatory Extremeness than those in the Web Search condition (WebSearch: M = - 0.53; p = 0.21). The difference was not statistically significant, which does not support \textbf{H6}. 

\subsection{STUDY 1: Result Summary}
In summary, Study 1 results showed that users of LLM-powered conversational search systems (ConvSearch and ConvSearchRef) exhibit higher levels of confirmatory information querying (H1) compared to users of conventional web search systems (WebSearch). Even with neutrally designed systems, we found evidence that LLM-powered conversational search systems led to higher degrees of opinion polarization regarding post-search perception of consonant versus dissonant information (H4 and partially H5), although we did not observe significant effects in self-reported confirmatory attitude change after the short search sessions nor differences across conditions in the confirmatory stances of participants' essays. 
\section{Study 2 Method: Effects of Opinionated LLM-Powered Conversational Search Systems}

Study 1 shows that, compared to conventional web search, conversational search, even when designed to be neutral, could lead to a higher level of confirmatory search behaviors and opinion polarization. Study 2 investigates whether a conversational search system powered by an LLM with an opinion bias can change these tendencies. Specifically, we ask whether a consonant LLM with an opinion bias that reinforces one's existing attitude exacerbates selective exposure; and whether a dissonant LLM with an opinion bias that challenges one's attitude mitigates selective exposure. 

We conducted another online experiment in Study 2. Like Study 1, participants were asked to perform an information-seeking task on a given controversial topic. We ran a 2x3 fully factorial between-subjects design where we compared two interfaces, ConvSearch and ConvSearchRef, and three opinion bias settings: Consonant, Neutral, and Dissonant. As we did not observe significant differences between ConvSearch and ConvSearchRef as in Study 1, we will focus on comparing the effects of the three different LLM opinion biases (i.e., aggregating the results with the two interfaces).

We adopted a largely similar study procedure, the same set of three topics, conversational search system UI, and measurements as in Study 1. Below, we will only discuss the additional step in Study 2 method: how we configured the opinionated LLMs to power the search systems, and the hypotheses. Study 2 was also approved by the Institutional Review Board of the author's institution. 

\begin{figure*}[t]
    \centering
    \includegraphics[width=0.95\textwidth]{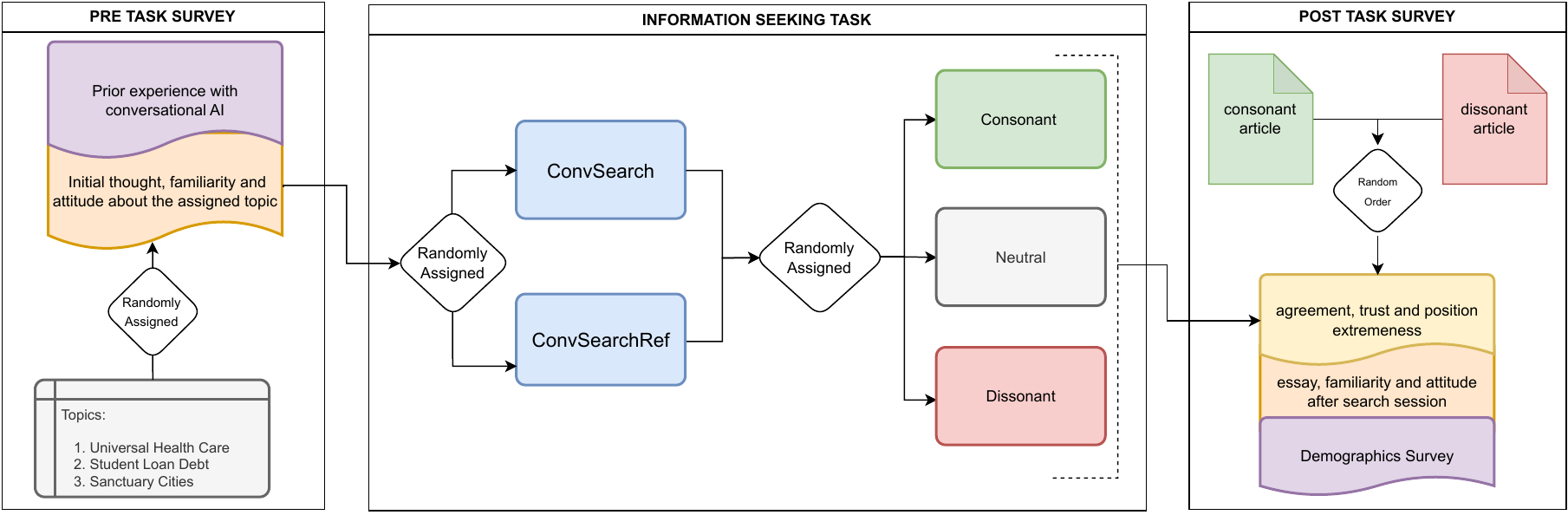}
    \caption{Overall study procedure for Study 2. In the pre-task survey, participants answered questions regarding their prior experience with conversational AI and their prior attitude and familiarity with a randomly assigned topic. Then, participants performed an information-seeking task to gather information on the topic with a randomly assigned information search system with a randomly assigned search system bias. After the search session, participants wrote an essay about the assigned topic. In the post-task survey, the participants again rated their attitude and familiarity with the topic, indicated their perception of two new articles on the topic (one consonant and one dissonant), and answered a demographic survey.}
    \label{fig:study2 procedure}
\end{figure*}

\subsection{Study Procedure}
As illustrated in Fig.~\ref{fig:study2 procedure}, the study procedure is largely similar to Study 1 (see Sec.~\ref{procedure_1}) with one exception: participants were randomly assigned to one of the six conditions as a combination of two conversational search system interfaces (ConvSearch or ConvSearchRef) and three manipulated opinion bias (Consonant, Neural, Dissonant). The biased LLM-powered conversational search system was assigned based on a participant's pre-existing attitude provided in the pre-task survey. That is, if a participant indicated they had an opposing attitude on the assigned topic, and they were assigned to be in the Consonant condition, then the system automatically selected the configuration with an opposing opinion bias on the topic. Similar to Study 1, participants were debriefed on the purpose of the study and their assigned conditions at the end, and provided with sources to a collection of articles that offered balanced and comprehensive information on the topic.

\subsection{Configuring Opinionated LLM-Powered Conversational Search Systems}
For each topic, we implemented two versions of biased conversational search systems in addition to the neutral one used in Study 1---one with a supporting bias and one with opposing bias towards the given controversial topic.

As shown in Fig. \ref{fig:CHAT_PROCESS}, we manipulated these opinion biases in the conversational search system with two modules in the RAG system architecture---information retrieval and response generation. For information retrieval, while the neutral system retrieves documents from a database with a balanced mixture of attitudes on the given topic (the same set of documents as used in Study 1), the biased system retrieves from documents of a biased database with only documents with the given bias and neutral documents. Note that the neutral documents often express balanced views so participants were still exposed to some different views even with a biased system.

For the response generation, we manually designed prompts to generate biased responses for the experiment. In pilot studies, we observed that an extremely strong system bias manipulation undermined the response quality generation, e.g., instead of answering people's queries, the system only expressed biased opinions on the topic, and the user quickly disengaged. We adjusted the prompts to balance between system bias and usability \footnote{\rr{See the Supplementary Material for system architecture. As discussed in Sec.~\ref{limitation}, prompts used in Study 2 will only be made available upon request to prevent misuse.} }

With the two biased (supporting/opposing) and the neutral configurations for each topic, we created three experimental conditions based on participants' pre-existing attitudes in the pre-task survey:
\begin{itemize}
    \item \textbf{Consonant}: The conversational search system is biased towards the participant's attitude, providing information that supports their pre-existing views. 
    \item \textbf{Neutral}: The conversational search system maintains a neutral stance, providing a balanced mixture of information from different viewpoints.  
    \item \textbf{Dissonant}: The conversational search system is biased against the participant's attitude, providing information that challenges their pre-existing views. 
\end{itemize}

\subsection{Hypotheses}
Our hypotheses are based on the same set of measurements as collected in Study 1 (see Sec.~\ref{measures}). An additional 265 search queries and 1000 essay sentences were coded with the same coding procedure in Study 1. In Study 2, we are interested in the possible different effects of consonant and dissonant conversational search systems. More specifically, we hypothesized that a consonant system may reinforce people's existing views and increase selective exposure, while a dissonant system may nudge people to seek diverse views and reduce opinion polarization. We made the following six hypotheses for the Consonant and Dissonant LLM-powered conversational search system, respectively.

\subsubsection{Hypotheses about Consonant Search System}
\begin{itemize}
    \item \textbf{[H1.a]}: When searching with a Consonant conversational search system, compared to a Neutral system, people will issue a \textit{higher} percentage of \textbf{Confirmatory Queries}.
    \item \textbf{[H2.a]}: When searching with a Consonant conversational search system,  compared to a Neutral system, people will exhibit a \textit{higher} level of \textbf{Confirmatory Attitude Change}. 
    \item \textbf{[H3.a]}: When searching with a Consonant conversational search system,  compared to a Neutral system, people will write a \textit{higher} percentage of \textbf{Confirmatory Argument} in their essays.
    \item \textbf{[H4.a]}: When searching with a Consonant conversational search system,  compared to a Neutral system, people will display a \textit{higher} level of \textbf{Confirmatory Agreement}.
    \item \textbf{[H5.a]}: When searching with a Consonant conversational search system,  compared to a Neutral system, people will display a \textit{higher} level of \textbf{Confirmatory Trust}.
    \item \textbf{[H6.a]}: When searching with a Consonant conversational search system,  compared to a Neutral system, people will display a \textit{lower} level of \textbf{Confirmatory Extremeness}.
\end{itemize}

\subsubsection{Hypotheses about Dissonant Search System}
\begin{itemize}
    \item \textbf{[H1.b]}: When searching with a Dissonant conversational search system,  compared to a Neutral system, people will issue a \textit{lower} percentage of \textbf{Confirmatory Queries}.
    \item \textbf{[H2.b]}: When searching with a Dissonant conversational search system,  compared to a Neutral system, people will exhibit a \textit{lower} level of \textbf{Confirmatory Attitude Change}.
    \item \textbf{[H3.b]}: When searching with a Dissonant conversational search system,  compared to a Neutral system, people will write a \textit{lower} percentage of \textbf{Confirmatory Argument} in their essays.
    \item \textbf{[H4.b]}: When searching with a Dissonant conversational search system,  compared to a Neutral system, people will display a \textit{lower} level of \textbf{Confirmatory Agreement}.
    \item \textbf{[H5.b]}: When searching with a Dissonant conversational search system,  compared to a Neutral system, people will display a \textit{lower} level of \textbf{Confirmatory Trust}.
    \item \textbf{[H6.b]}: When searching with a Dissonant conversational search system,  compared to a Neutral system, people will display a \textit{higher} level of \textbf{Confirmatory Extremeness}.

\end{itemize}

\begin{figure*}[t!]
    \centering
    \includegraphics[width=0.8\textwidth]{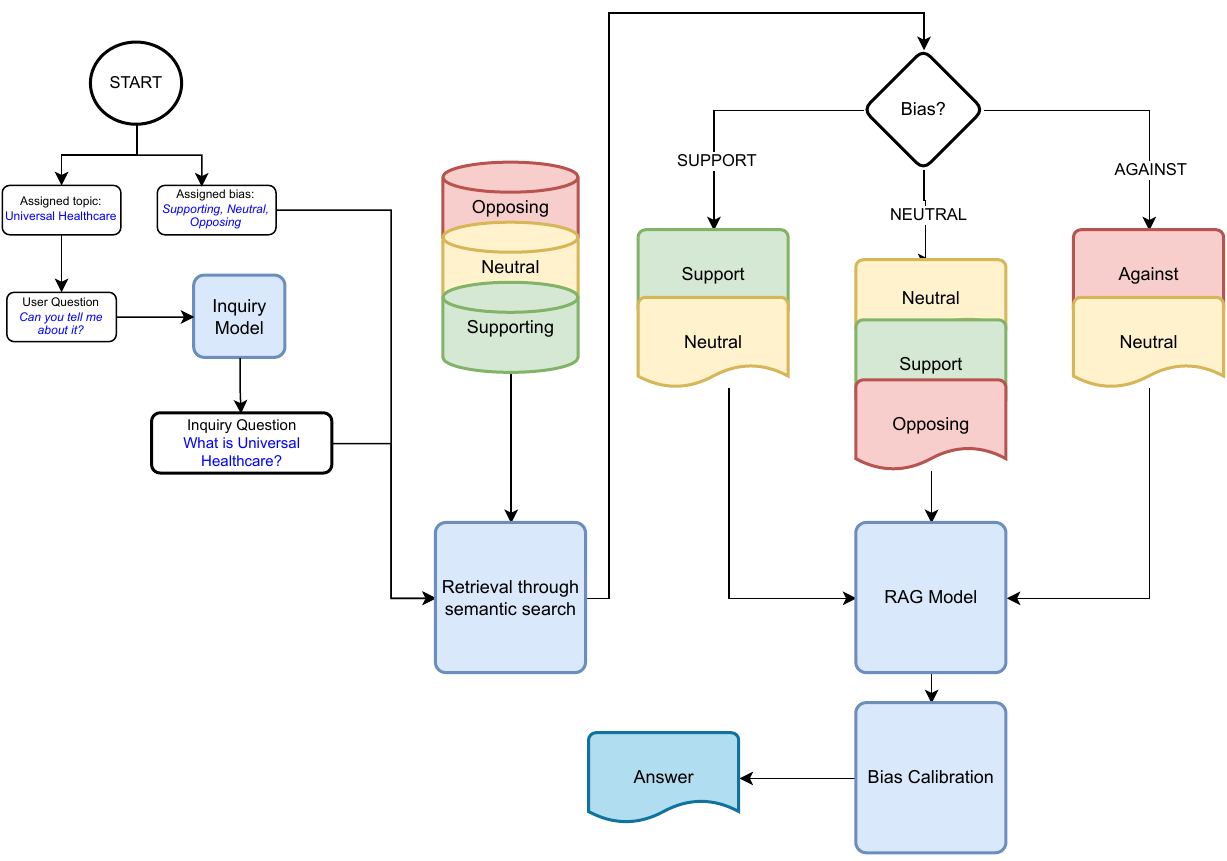}
    \caption{System Architecture of the \textit{Opinionated} LLM-powered Conversational Search system. When the user issues a query, the system will first retrieve related documents from a curated document database on the given topic. By adjusting the bias mixture of the document pool, along with a set of handcrafted prompts, the system will produce a response that is either consonant with the user's attitude, dissonant with the user's attitude, or neutral. The backbone model of our system is gpt-4-32k-0613.}
    \label{fig:CHAT_PROCESS_study2}
\end{figure*} 

\subsection{Analysis Plan}
 We again ran the analysis of covariance (ANCOVA) with Tukey's method (p-values adjusted for multiple comparisons) to conduct post-hoc analysis when ANCOVA showed significance. In each ANCOVA analysis, the independent variable was the search system bias (Consonant, Neutral, and Dissonant), and the dependent variable was a measure in a hypothesis. Control variables include search interface, participants' demographics, their prior experience with conversational AI, usage frequency, assigned topic, and participant's pre-existing attitudes to the topic.
All analysis results and descriptive statistics are listed in Tab.~\ref{tab:overall-study2}. When discussing the results below, we will focus on highlighting the patterns.

\subsection{Participant Overview}

In addition to the participants in ConvSearch and ConvSearchRef condition in Study 1 (Neutral condition), we recruited an additional 148 participants to interact with opinionated conversational search systems. Similar to Study 1, we recruited fluent English speakers from the United States on Prolific. The compensation rate was \$15 per hour. Participants in Study 1 were not allowed to participate in Study 2. As a result, Study 2 included 223 participants (Tab.~\ref{tab:participants-study2}). Among them, 112 identified as women, 104 identified as men, and 7 identified as non-binary or third gender. Similar to participants in Study 1, the median education level was a Bachelor's degree. The median household income was between \$50,000 - \$ 100,000. The median age of participants was between 25 and 34 years old. 

\begin{table}[b!]
\vspace{-12mm}
\small
\begin{tabular}{lllll}
\hline
              & \textbf{Consonant} & \textbf{Neutral} & \textbf{Dissonant} & Total \\ \hline
ConvSearch    & 37                 & 38               & 38                 & 113   \\
ConvSearchRef & 36                 & 37               & 37                 & 110   \\ \hline
Total         & 73                 & 75               & 75                 & 223  
\end{tabular}
\vspace{8mm}
\caption{Participants distribution in Study 2. 73 participants interacted with a conversational search system aligned with their pre-existing attitude on the topic, 75 participants interacted with a neutral system, and 75 participants interacted with a system biased against their pre-existing attitude.}
\label{tab:participants-study2}
\end{table}

\section{Study 2 Results}
\subsection{Manipulation Checks}
We performed manipulation checks on the perceived bias (0: Dissonant, 3: Neutral, 5: Consonant) of the opinionated conversational search systems, participants in the Consonant condition believed the system was biased toward them (M = 3.28, SD = 0.68), participants in the Neutral condition did not perceive any bias from the search system (M = 3.05, SD = 0.57), and participants in the Dissonant conditions perceived the search system is against their attitude (M = 2.64, SD = 0.86). An ANCOVA analysis showed there was a significant difference across conditions (F(2, 210) = 15.53, p < 0.001 *). Post-hoc analysis showed the differences between Consonant and Dissonant (p < 0.001 ***) and between Neutral and Dissonant (p = 0.001 **) were significant. Interestingly, the perceived difference between Consonant and Neutral was not significant (p = 0.13).

The second manipulation check looks at the effectiveness of the search session. There was a significant self-report familiarity change (Pre-Search: Mean = 3.31, SD = 1.13; Post-Search: Mean = 3.85, SD = 0.87) and no significant difference across conditions (Consonant: M = 0.63, SD = 1.02; Neutral: M = 0.41, SD = 0.93; Dissonant: M = 0.57, SD = 0.92; F(2, 200) = 1.06, p = 0.35). These results indicated that in all conditions, participants reported searching with the system made them more familiar with the given topic.

On average, our participants spent 20.63 mins (SD = 10.29) completing the study. Although participants in the Consonant (M = 21.15 mins, SD = 10.60) or Dissonant (M = 21.15 mins, SD = 9.98) spent a longer time than the Neutral condition (M = 19.61 mins, SD = 10.35), the difference was not significant. Participants, on average, issued 3.69 queries per search session (SD = 1.10). An ANCOVA analysis did not suggest differences in terms of search conditions (Consonant: M = 3.65, SD = 0.96; Neutral: M = 3.53, SD = 0.91; Dissonant: M = 3.88, SD = 1.35; F(2, 210) = 1.90, p = 0.15).

\begin{table*}[t]
\begin{center}
\resizebox{\linewidth}{!}{
    \begin{tabular}{cccccc } 
     \toprule
     \textbf{Hypothesis} 
&    $\begin{matrix}\textbf{Consonant} \end{matrix}$ 
&    $\begin{matrix}\textbf{Neutral} \end{matrix}$ 
&    $\begin{matrix}\textbf{Dissonant} \end{matrix}$ 
& \textbf{Post-Hoc Analysis} \\
     \midrule
     $\begin{matrix}
        \textbf{H1: Confirmatory Query}
        \\
        F(2,210) = 31.24, p < 0.001\text{***}
     \end{matrix}
     $ 
& $\begin{matrix}Mean = 42.92\% \\ SD = 29.11\% \end{matrix}
     $& $\begin{matrix}Mean = 15.57\% \\ SD = 29.11\% \end{matrix}
     $ & $\begin{matrix}Mean = 12.33\% \\ SD = 24.60\% \end{matrix}
     $ & 
     $\begin{matrix}
     \text{Consonant > Neutral ***}
     \\
     \text{Consonant > Dissonant ***}
     \end{matrix}
     $

     \\     \midrule
     $\begin{matrix}
     \textbf{H2: Attitude Change}
     \\
      F(2,210) = 3.36, p = 0.04\text{*}  
     \end{matrix}$    
     & $\begin{matrix}Mean = 0.27 \\ SD = 0.75 \end{matrix}
     $& $\begin{matrix}Mean = 0.00 \\ SD = 0.65 \end{matrix}
     $ & $\begin{matrix}Mean = 0.08 \\ SD = 0.73 \end{matrix}
     $& $\begin{matrix}\text{Consonant > Neutral}\dagger\end{matrix}
     $
  \\           \midrule
    $\begin{matrix}
    \textbf{H3: Confirmatory Argument}
    \\
    F(2,210) = 10.30, p < 0.001\text{***}
    \end{matrix}
     $ 
& $\begin{matrix}Mean = 51.69\% \\ SD = 43.00\% \end{matrix}
     $& $\begin{matrix}Mean = 34.79\% \\ SD = 48.76\% \end{matrix}
     $ & $\begin{matrix}Mean = 15.58\% \\ SD = 58.18\% \end{matrix}
     $ &  
     $\begin{matrix}
     \text{Consonant > Dissonant ***} \\
     \text{Consonant > Neutral}\dagger
     \\\text{Neutral > Dissonant}\dagger
     \end{matrix}
     $
     \\
          \midrule
     $\begin{matrix}
     \textbf{H4: Confirmatory Agreement}
     \\
     F(2,210) = 7.43, p < 0.001\text{***}
     \end{matrix}$
     & $\begin{matrix}Mean = 2.44 \\ SD = 1.27 \end{matrix}
     $& $\begin{matrix}Mean = 1.84 \\ SD = 1.55 \end{matrix}
     $ & $\begin{matrix}Mean = 1.51 \\ SD = 1.75 \end{matrix}
     $&  
     $\begin{matrix}
     \text{Consonant > Dissonant ***} \\
     \text{Consonant > Neutral *}
     \end{matrix}
     $
     \\
    \midrule
    $\begin{matrix}
    \textbf{H5: Confirmatory Trust}
    \\
    F(2,210) = 2.91, p = 0.057 \dagger
    \end{matrix}$
     & $\begin{matrix}Mean = 1.24 \\ SD = 1.02 \end{matrix}
     $& $\begin{matrix}Mean = 0.84 \\ SD = 0.97 \end{matrix}
     $ & $\begin{matrix}Mean = 0.93 \\ SD = 1.24 \end{matrix}
     $
& 
     $\begin{matrix}
     \text{Consonant > Dissonant}\dagger \\
     \text{Consonant > Neutral}\dagger
     \end{matrix}
     $
     \\     
      \midrule
     $\begin{matrix}
     \textbf{H6: Confirmatory Extremeness}
     \\ 
     F(2, 210) = 3.30, p = 0.04\text{*}
     \end{matrix}$
     & $\begin{matrix}Mean = -1.63 \\ SD = 1.49 \end{matrix}
     $& $\begin{matrix}Mean = -1.09 \\ SD = 1.67 \end{matrix}
     $ & $\begin{matrix}Mean = -1.03 \\ SD = 1.52 \end{matrix}
     $ &      $\begin{matrix}
     \text{Consonant < Dissonant}\dagger \\
     \text{Consonant < Neutral}\dagger
     \end{matrix}
     $
    \\      
    \end{tabular}

}
\end{center}
\vspace{3mm}
\caption{Summary of quantitative results from Study 2. The left column shows $p$-values obtained via ANCOVA tests for each hypothesis. The right column shows pairs of conditions that are statistically significantly different or marginally significant. Significance is marked as 
$p < 0.1$ ($\dagger$), $p < 0.05$ (*), $p < 0.01$ (**), or $p < 0.001$ (***).}
\label{tab:overall-study2}
\end{table*}

\subsection{Consonant Conversational Search Induced Higher Level of Confirmatory Information Seeking (H1.a Supported; H1.b Not Supported)}

We found that participants in the Consonant condition issued more confirmatory queries (Consonant: M = 42.92 \%) than their counterparts did in Neutral and Dissonant conditions (Neutral: M = 15.57 \%; Dissonant: M = 12.33 \%; p < 0.001 ***), details see Tab.~\ref{tab:overall-study2}. The pair-wise comparison showed that the difference was significant between Consonant and Dissonant (p < 0.001 ***; Cohen's D = 1.19) and between Consonant and Neutral (p < 0.001 ***; Cohen's D = 1.01); both indicated large effect sizes. 

This result supports \textbf{H1.a} that a Consonant conversational search system leads to more confirmatory information-seeking behaviors compared to a Neutral system, suggesting that people's confirmatory information-seeking behaviors can be further biased when having conversational interactions that reinforce their existing views.  However, we found no evidence supporting \textbf{H1.b}, which indicates a limited effect of using Dissonant conversational search to nudge people towards more diverse information-seeking behaviors. 

\subsection{Consonant Conversational Search Induced Higher-level of Opinion Polarization (H2.a-6.a Mostly Supported; H2.b-6b Mostly Not Supported)}
\subsubsection{Participant's Confirmatory Attitude Change}
We observed differences in participants' self-reported confirmatory attitude change after the search session (Consonant: M = 0.27; Dissonant: M = 0.08; Neutral: M = 0.00). The ANCOVA analysis showed significance (p = 0.04 *). Post-Hoc analysis showed that the difference between the Consonant and Neutral conditions is marginally significant (p = 0.053, Cohen's D = 0.39). The result partially supports \textbf{H2.a} showing that searching with a Consonant system could lead to more polarized attitude change with even a short search session. In contrast, there is no support for \textbf{H2.b}) with regard to the effect of using a Dissonant system.

\subsubsection{Confirmatory Arguments} 
The results showed that an opinionated LLM-powered conversational search system could skew the content people wrote in their essays. We found significant differences in Confirmatory arguments in the final essays across conditions (Consonant: M = 51.69 \%; Neutral: M = 34.79 \%; Dissonant: M = 15.58 \%; p < 0.001 ***). Post-Hoc analysis showed marginal differences between Consonant and Neutral (p = 0.09, Cohen's D = 0.7) and between Neutral and Dissonant (\textbf{H3.b}: p = 0.05, Cohen's D = 0.36). The results provide partial support for \textbf{H3.a} and \textbf{H3.b}, suggesting that interacting with the Consonant system led to higher opinion polarization; meanwhile, the Dissonant system has the potential to reduce opinion polarization, at least regarding the information people produce after the search session. 

\subsubsection{Confirmatory Perception of Given Articles} Similar to Study 1, we asked participants to rate three types of perceptions of a consonant article and a dissonant article after the search session to measure opinion polarization: agreement, trust, and extremeness. 

For agreement, participants agreed with the consonant article (M = 4.07, SD = 0.80) and disagreed with the dissonant article (M = 2.15, SD = 1.12). The ANCOVA analysis showed that participants in the Consonant condition (M = 2.44) displayed a significantly higher level of Confirmatory Agreement than those in the Dissonant condition (M = 1.51) and Neutral condition (M = 1.84; p < 0.001 ***). Post-Hoc analysis found the differences between Consonant and Dissonant (p < 0.001 ***, Cohen's D = 0.61) and between Consonant and Neutral (p = 0.04 *, Cohen's D = 0.42) were statistically significant. The difference between Dissonant and Neutral was not significant (p = 0.38). The results support \textbf{H4.a} but not \textbf{H4.b}.

For trust, participants trusted the consonant (M = 4.01, SD = 0.74) more than the dissonant article (M = 3.00, SD = 0.81). The Consonant search system led to a higher level of Confirmatory Trust (Consonant: M = 1.24) than the other two systems (Neutral: M = 0.84; Dissonant: M = 0.93), with the ANCOVA test showing marginal significance (p = 0.057). The differences between Consonant and Neutral (p = 0.094, Cohen's D = 0.41) and between Consonant and Dissonant (p = 0.05, Cohen's D = 0.27) were marginally significant. The difference between Dissonant and Neutral was not significant (p = 0.96). The results partially support \textbf{H5.a} but not for \textbf{H5.b}. 

For extremeness, participants perceived the dissonant article as more extreme (M = 3.58, SD = 1.05) than the consonant article (M = 2.33, SD = 1.02). Participants who searched with the Consonant system exhibited a lower level of Confirmatory Extremeness (Consonant: M = -1.63) than participants in the other two conditions (Neutral: M = -1.09; Dissonant: M = -1.03; p = 0.04 *). In the Post-Hoc analysis, the results showed that the differences between Consonant and Dissonant conditions (p = 0.05, Cohen's D = -0.40) and between Consonant and Neutral (p = 0.09, Cohen's D = -0.34) were marginally significant, but not between Neutral and Dissonant conditions (p = 0.96). \textbf{H6.a} is partially supported but \textbf{H6.b} is not.

\subsection{STUDY 2: Result Summary}
In conclusion, Study 2 revealed that opinionated LLM-powered conversational search systems can significantly influence people's information-seeking behaviors and opinions, and whether the encoded opinion bias was consonant or dissonant with people's existing views had distinct effects. Participants interacting with a Consonant system exhibited more confirmatory queries (\textbf{H1.a}), and a significantly higher degree of opinion polarization across all measures (\textbf{H2.a}-\textbf{H6.a}). In contrast, we found that interacting with a dissonant system had a rather limited effect in mitigating confirmatory information-seeking and opinion polarization. These findings highlight the potential risks associated with opinionated LLM-powered search systems in reinforcing people's existing beliefs and biases.
\section{Discussion}
Through two controlled experiments, we demonstrate the risks of LLM-powered conversational search in exacerbating people's selective exposure bias and opinion polarization. We found that, even with a neutral LLM-powered search system (regardless of whether source references are provided or not), participants exhibited significantly more bias of their pre-existing views in their information queries compared to when using a conventional web search system. This biased querying behavior led to some degree of opinion polarization regarding the post-search perception of consonant versus dissonant information, which risks further skewing people's future information consumption. This bias towards seeking consonant information was even more pronounced when using a conversational search system powered by an opinionated LLM that reinforces participants' pre-existing views, leading to significantly more opinion polarization across all measures compared to when using a neutral LLM-powered conversational search system. Interestingly and alarmingly, interacting with a dissonant LLM-powered conversational search system with the opposite opinion had little effect in reducing the selective exposure bias in information querying and opinion polarization (with the exception of a more balanced view in participants' essays). Below, we interpret the potential mechanisms, suggest strategies to mitigate the echo chamber effect in conversational search, and consider our results' implications for potential harms brought by LLMs.

\subsection{Selective Information Seeking with Conversational Search}
\label{mechanism}
Our results suggest that the natural conversational interactions enabled by LLMs exhibit more of people's existing biases, and more so when the LLMs have reinforcing opinion biases. There can be multiple mechanisms contributing to this phenomenon. First, consistent with prior work studying a spoken conversational search system~\cite{trippas2018informing}, we observed that compared to keyword-based search, participants' conversational queries were more verbose and expressive. For example, one participant asked \textit{``College here in the USA is disgusting overpriced and greedy. Wouldn't it be better to look at that as the issue instead of keeping our current greedy practices and debating about forgiving some?''}. It is also possible that conversational interactions resemble social interactions, and people are more likely to engage in opinionated communication, especially when the other party reinforces their views (i.e., consonant LLM). For example, one asked \textit{``Yeah, give me that information please. Tell me about the arguments in favor of sanctuary cities.''}. Linguistic and communication accommodation~\cite{giles1991accomodation,danescu2011mark}, with which people converge to the conversational partners' communication behaviors, could also have played a role. 

We must note that there may exist additional differences in information consumption mechanisms besides the difference in querying behavior when using conversational search versus conventional search systems. These mechanisms may bring in different selective exposure biases. With a conventional search system, people engage in additional information selection through \textit{clicks} of links. Indeed, we observed that in the Web Search Condition, on average, participants clicked 4.48 (SD = 4.27) links of consonant articles versus 3.28 (SD = 3.11) dissonant articles. In theory, a neutral LLM-powered conversational search would synthesize the retrieved articles in a relatively faithful fashion (i.e., reflecting the overall position of the retrieved articles). However, people may place selective attention or retention on these synthesized outputs. Indeed, we observed that participants spent an average of 116.6 seconds (SD = 177.62) reading consonant outputs versus 78.66 seconds (SD = 111.90) reading dissonant outputs in the two conversational search conditions; they also gave 0.29 thumbs-up (and 0.08 thumbs-down) to consonant outputs versus 0.21 (and 0.04 thumbs-down) to dissonant outputs.

While we observed these additional behavioral biases, our study does not fully capture participants' information consumption patterns, nor provide comparable ways to characterize how these patterns impact attitude polarization in conversational and conventional search differently. We encourage future work to explore these questions empirically (e.g., through eye-tracking studies) and develop a more principled understanding of where and how cognitive biases could impact people's information consumption using conversational search systems.

\subsection{Mitigating Selective Exposure in Conversational Search}
Surprisingly, our results suggest that injecting the opposite opinion bias in LLM-powered conversational search systems may have a limited effect in combating selective exposure. It is, therefore, necessary to resort to other design interventions to mitigate selective exposure and increase people's information diversity. HCI research has a long history of developing diversity-enhancing designs and systems~\cite{kriplean2012supporting,faridani2010opinion,nagulendra2014understanding,jeon2021chamberbreaker}, ranging from deliberation platforms, visualization systems, and diverse news feeds. In particular, Liao et al.~\cite{liao2014can,liao2014expert,liao2015all} draw lessons from psychology research on selective exposure~\cite{hart2009feeling} and recommend targeting two fundamental psychological mechanisms that can reduce individuals' selective exposure: increasing people's accuracy motivation to learn accurate and comprehensive information, and/or reducing people's defense mechanism when being confronted with opposing views. Example designs that can increase accuracy motivation include highlighting the values of the information with opposing views, such as the expertise of the information source or new knowledge it brings, or making people aware of their own bias. A conversational search system can leverage simple nudges through conversations, such as reminding the user of biases in the information they consumed and suggesting diverse queries to try. Example designs that can decrease defense mechanisms include acknowledging the common ground in the opposing views, and presenting diverse perspectives with agreeable information to make them easier to consume. Future work should explore leveraging these more sophisticated communication and presentation strategies in LLM outputs.

It is worth noting that the search systems we tested constitute an active information-seeking paradigm, and the information consumption is predominantly determined by the people' querying behaviors. People also receive information passively or through scanning. Indeed, prior works argued that conversational search and agent systems have the potential advantage of making individuals more receptive toward proactive interactions from the system~\cite{radlinski2017theoretical}, and HCI researchers explored leveraging conversational systems such as Alexa to broadcast diverse views~\cite{feltwell2020broadening}. It would be interesting to explore whether conversational interactions enabled by LLMs can effectively act as active nudging for diverse views, and whether they provide benefits over conventional information systems such as news feed and deliberation platforms. 

\subsection{Effect of References in LLM-powered Information Systems}
While earlier LLM-powered information systems such as ChatGPT often generate responses to a user's question without specifying the information sources, most recent LLM-powered search systems, including Bing Chat and Google Bard, added the source reference feature as essential for ensuring information credibility for the search experience. However, both of our studies showed that including references had a very limited impact on people's information-seeking behavior and opinion polarization. 

We observed that, on average, participants clicked less than one reference per search session (M = 0.43, SD = 1.13). This implies people's low willingness to engage with the information source feature in today's LLM-powered conversational search system (though admittedly, participants did not necessarily perform a high-stake task in our study). Such low engagement may put people at risk especially given current LLMs' tendency to generate non-factual information. For example, a recent study \cite{liu2023evaluating} found that popular commercial LLM-powered search systems frequently output inaccurate information and wrong references. We encourage future research to explore designs that guide people to verify information generated by LLMs and attend to the information sources. The reference feature can also be leveraged to provide additional opportunities for exposure to diverse information, such as encouraging people to check out sources that present balanced views.

\subsection{Implications for Information Harms of LLMs}
LLMs have already reached hundreds of millions of users and their impact is poised to continue growing. Many are concerned about the potential harms they can bring to individuals and society. A unique aspect of LLMs is that they produce content and information, which can be consumed and circulated, impacting multiple stakeholder parties including the co-creator of the information, the readers of the information, as well as subjects described in the texts~\cite{rauh2022characteristics}. These processes can lead to multiple types of harm~\cite{shelby2022sociotechnical,weidinger2022taxonomy}, from discrimination and exclusion, to disinformation and misinformation, as well as creating information hazards such as leaking sensitive information or compromising privacy. Our results further imply that the conversational interaction affordance of LLMs may have a reinforcing effect on their \textit{information harms}, and are especially likely to harm those already with a predisposition towards the biases, misinformation, disinformation or other information hazards produced by LLMs. In other words, potential harms of LLMs should be approached as sociotechnical problems ~\cite{liao2023rethinking}, considering not only the limits of the technology, but also people's interaction behaviors with specific LLM-powered applications. 

As large-scale information and knowledge systems, we must consider LLMs' societal risks through ``subjugation''~\cite{shelby2022sociotechnical}---how they can proliferate dominant views and languages and foreclose alternative ones. A previous study by~\citet{jakesch2023co} highlights that LLM-powered writing support can exert latent persuasion with biases in its generated content. Our results imply an additional mechanism for such a risk by creating echo chambers. We must consider the risks from both dominant views and biases encoded in widely used LLMs, as well as the danger of targeted opinion influence by political or commercial groups that exploit the echo chamber effect. As our Study 2 shows, it can be extremely easy to steer LLMs to exhibit certain biases through adaptation techniques. Extra attention is demanded as LLMs in information systems often serve multiple functions (e.g., retrieval and response generation). Without proper auditing and mitigation, biases encoded in an LLM can slip in to produce unwanted information risk. These biased LLMs can be used for not only conversational search but also writing support, chatbots, social media bots, and so on. Overall, our results suggest that opinion biases in LLMs present risks that may outweigh the potential benefits. We encourage the development of technical guardrails and auditing methods to detect opinion biases of LLMs and prevent malicious manipulation of such biases. Policymakers and society at large must also grapple with how to establish norms and regulations to restrain such manipulation and make transparent LLMs' possible opinion biases. We also encourage future work to explore how the embedding of LLM-powered information systems and agents can shape public opinions.    

\subsection{Limitations and Ethical Considerations}
\label{limitation}
We acknowledge several limitations of our study. First, we adopted a closed-world version of search systems and a highly focused essay-writing task. The results may not be fully generalizable to diverse real-world information-seeking settings with search systems. Second, we acknowledge that there can be alternative implementations of LLM-powered conversational search systems and ways to create opinionated LLMs, and they may exhibit different biases and influences. For example, the RAG architecture utilizes prompt engineering to ensure the synthesized outputs to reflect the views in the retrieved articles in a relatively faithful and balanced fashion. It is possible that without this layer, LLMs are more likely to generate more catering responses that can introduce further biases. In Study 2, we manipulated the LLM's bias in both the retrieval and the response generation modules. This approach may create opinionated search systems with strong biases, which may not be as pronounced in other settings especially when biases are ``unintended''.  Second, participants engaged in relatively short search sessions (an average of 3.40 queries per session) with a specific task of essay writing. The results may not be generalizable to other information-seeking settings that require longer or continuous learning. However, we emphasize that differences in querying behavioral patterns and opinion polarization can be observed even in such short interaction sessions. Third, as discussed in Section~\ref{mechanism}, our study focuses on information querying behaviors without capturing other information consumption mechanisms such as selective attention, perception, and retention that can further affect opinion polarization. Lastly, it is known that there are individual differences in diversity-seeking tendencies~\cite{liao2013beyond,munson2010presenting} and tendencies to engage in human-like conversational interactions with machines~\cite{liao2016can}. Our study only looked at participants' behaviors at an aggregate level without considering these individual differences.

We also acknowledge that our results and system design may incur misuse. Overall, we caution against using LLMs with opinion biases to power search systems without appropriate risk assessment and oversight, which should be enforced not only through technical guardrails but also norms and policy. We hope our results can inform such guardrails. To prevent misuse, we decided not to make public the prompts we used to generate biased LLM responses but will only make them available for requests that we can verify for safe usage (e.g., scientific and non-commercial purposes).
\section{Conclusion}
The recently developed powerful LLMs have experienced exponential growth in user population. They have been adopted to power numerous information systems, such as conversational search, open-domain, or specialized chatbots, to various productivity tools. These applications can have a profound impact on the ways people search and consume information. In this study, through two controlled experiments, we empirically showed: 1) LLM-powered conversational search could lead to increased selective exposure and opinion polarization compared to conventional web search, by inducing more confirmatory querying behaviors in conversational interactions; 2) an opinionated LLM that reinforces the user's view could exacerbate the effect, together suggesting the risk of ``generative echo chambers''. Our study also suggests the limitations of interventions such as providing references and leveraging an LLM that challenges one's existing view, both of which had little effect in reducing selective exposure. With millions of people already being exposed to LLM-powered information technologies, these results call for actions to regulate the use of LLM-powered search systems, develop technical guardrails against misuses of LLMs for opinion influence, and explore mitigation strategies for selective exposure in conversational search.
\begin{acks}
We thank Susan Dumais for providing thoughtful feedback. We also thank our participants for their participation and anonymous reviewers for their constructive comments on this work.
\end{acks}

\bibliographystyle{ACM-Reference-Format}
\bibliography{main}

\appendix
\end{document}